\documentclass[11pt,a4paper]{article}
\usepackage[hyperref]{acl2021}
\usepackage{latexsym}
\usepackage{times}
\usepackage{helvet}
\usepackage{courier}
\usepackage{todonotes}
\usepackage{amsmath}
\usepackage{url}
\usepackage{caption}
\usepackage{subfigure}
\usepackage{svg}
\usepackage{graphicx}
\usepackage{bm}
\usepackage{amssymb}
\usepackage{epstopdf}
\usepackage{scrextend}
\usepackage{multirow}

\aclfinalcopy 
\def\aclpaperid{4181} 

\title{SemGloVe: Semantic Co-occurrences for GloVe from BERT}

\author{
\textbf{Leilei Gan\textsuperscript{\rm 1},
Zhiyang Teng\textsuperscript{\rm2},
Yue Zhang\textsuperscript{\rm 2}},\\

\textbf{Linchao Zhu\textsuperscript{\rm 3},
Fei Wu\textsuperscript{\rm 1},
Yi Yang\textsuperscript{\rm 1}}\\

\textsuperscript{1} College of Computer Science and Technology, Zhejiang University \\
\textsuperscript{2} Westlake University,
\textsuperscript{3} University of Technology Sydney\\

{\tt \{leileigan, wufei, yangyics\}@zju.edu.cn},\\
{\tt \{tengzhiyang, zhangyue\}@westlake.edu.cn},\\
{\tt linchao.zhu@uts.edu.au}
}

\date{}

\begin{document}

\maketitle

\begin{abstract}
GloVe learns word embeddings by leveraging 
statistical information from word co-occurrence matrices. However, word pairs in the matrices are extracted from a predefined local context window, which might lead to limited word pairs and potentially semantic irrelevant word pairs. In this paper, we propose \emph{SemGloVe}, which distills \emph{semantic co-occurrences} from BERT into static GloVe word embeddings. Particularly, we propose two models to extract co-occurrence statistics  based on either the masked language model or the  multi-head attention weights of BERT. Our methods can extract word pairs limited by the local window assumption, and can define the co-occurrence weights by directly considering the semantic distance between word pairs. Experiments on several word similarity datasets and  external tasks show that SemGloVe can outperform GloVe.
\end{abstract}

\section{Introduction} 
\label{introduction}
Word embeddings \cite{bengio2003neural,mikolov2013efficient,mikolov2013distributed} represent words with low-dimensional real value vectors. They can be useful for lexical semantics tasks, such as word similarity and word analogy, and downstream natural language processing (NLP) tasks \cite{collobert2011natural,rajpurkar2016squad,lewis2016lstm,lee2018higher}. Most existing methods use local window-based methods \cite{mikolov2013efficient,mikolov2013distributed,bojanowski2017enriching} or matrix factorization of global statistics \cite{pennington2014glove} to learn syntax and semantic information from large-scale corpus.  In this paper, we investigate GloVe in details.

\begin{figure}[t]
    \centering
    \includegraphics[width=0.5\textwidth]{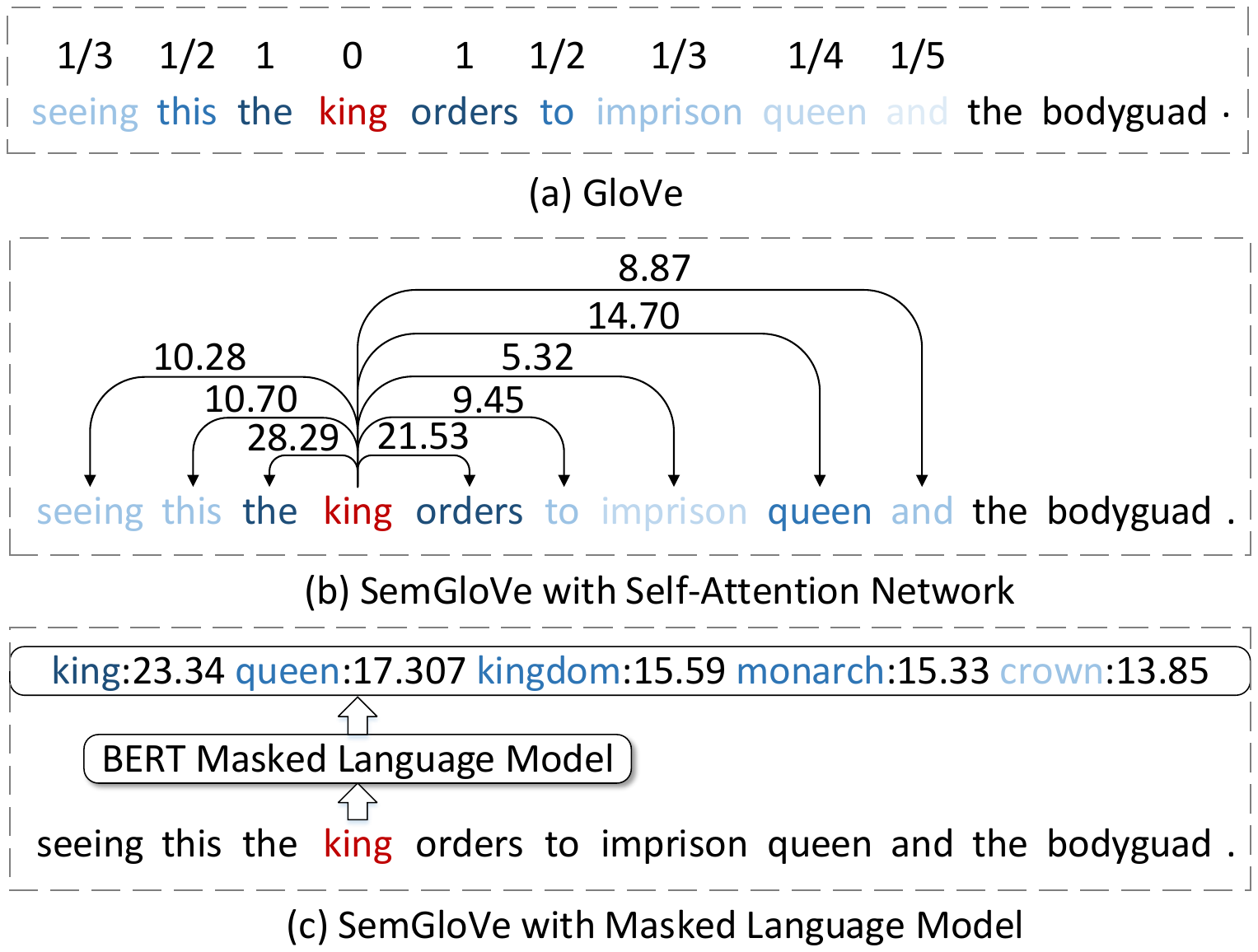}
	\caption{Glove (a) and Our two proposed models (b) and (c). The target word and the context word are in red and blue colors, respectively. In (a), the numbers are relevance score, which are position-based distance. In (b) and (c), the relevance scores are the self-attention weights, and the logits of the language model, respectively.}
	\label{Model Architecture}
\end{figure}

GloVe \cite{pennington2014glove} combines matrix factorization methods with local window context, generating word embeddings by leveraging statistical information from a global word-word co-occurrence matrix. Word-word pairs in the matrix are extracted from a predefined local context window, and the relevance is measured by a position-based distance function. For example, as shown in Figure \ref{Model Architecture} (a), when targeting at the word ``king'' with the five-words context window, the word pair ``king-queen'' can be extracted with a co-occurrence score 1/4, since ``queen'' is four words away from ``king''. 
However, this co-occurrence matrix generation procedure can suffer from two potential problems.
First, the counting word pairs are limited by the local context window.
Second, the heuristic weighting function does not measure the relevance score directly regarding to the semantic similarities between word pairs, leading to inaccurate co-occurrence counts. For example, ``king'' and ``orders'' are weakly correlated in the sentence ``he orders a copy of the king of fighters".

One potential solution to the above problems is BERT \cite{devlin2019bert}, which is a pre-trained language model based on deep bidirectional Transformers \cite{vaswani2017attention}. Previous work has shown that the contextualized representations produced by BERT capture morphological \cite{peters2018dissecting,hewitt2019structural}, lexical \cite{peters2018dissecting}, syntactic \cite{li-eisner-2019-specializing,goldberg2019assessing} and semantic knowledge \cite{da-kasai-2019-cracking,ethayarajh-2019-contextual}. These knowledge can be disentangled using knowledge distillation models \cite{hinton2015distilling,tang2019distilling} or variational inference \cite{li-eisner-2019-specializing}. Inspired by these research ideas, we hypothesize that the word-word co-occurrence matrix can be distilled from BERT.

We name the co-occurrences distilled from BERT as \emph{semantic co-occurrences}, thus proposing \emph{SemGloVe} to improve GloVe. Our proposed method can distill contextualized semantic information from BERT into static word embeddings. In particular, we present two models. As shown in Figure \ref{Model Architecture} (b), the first model is based on self-attention networks of BERT (SAN; Section \ref{sec:san}). The idea is to use attention values in SAN for semantic co-occurrence counts. Intuitively, the SAN model can solve the second issue of GloVe by using the self-attention weights as the word pair scoring function. For example, the relevance score of the ``king-queen'' pair in Figure \ref{Model Architecture} (b) is 14.7, which is much larger than the 9.45 score of the ``king-to'' pair even though ``to'' is closer to ``king'' than ``queen'' is.

As shown in Figure \ref{Model Architecture} (c), the second is based on the masked language model (MLM; Section \ref{sec:mlm}) of BERT. The idea is to distill word probabilities from a masked language model for co-occurrence counts. First, it generates word pairs by masking the target word to predict context words from the whole vocabulary, which can avoid the local context window restriction of GloVe. The output context words and the target word can be regarded as co-occurring word pairs. For example, in Figure \ref{Model Architecture} (c), after masking ``king'', BERT outputs context words such as ``queen'' and ``crown''. Both ``king-queen'' and ``king-crown'' are regarded as valid co-occurrences even though ``crown'' does not appear in this sentence. 
Compared with the local window context, we also hypothesize that the distributional hypothesis \cite{harris1954distributional} works better for the masked language model, because words that occur in the same output contexts tend to have similar meanings. Second, the MLM model uses the logits of output words from BERT as the word pair scoring function, which can solve the first problem of GloVe.


Experiments on word similarity datasets show that SemGloVe can outperform GloVe. We also evaluate SemGloVe on external Chunking, POS tagging and named entity recognition (NER) tasks, again showing the effectiveness of SemGloVe. Specifically, our analysis shows that SemGloVe has two advantages: first, it can find semantic relevant word pairs which cannot be captured by GloVe; second, SemGloVe can generate more accurate global word-word co-occurrence counts compared to GloVe. 

SemGloVe gives better averaged results compared with existing state-of-the-art (non-contextualized) embedding methods on both intrinsic and extrinsic evaluation tasks. In addition to its theoretical interest, this leads to two contributions to the research community. First, it enriches the toolbox for computational linguistics research involving word representations that do not vary by the sentence, such as lexical semantics tasks. Second, it adds to the set of input embeddings that are orders of magnitudes faster compared with contextualized embeddings. Our code and SemGloVe embeddings will be made available.

\section{Related Work}
\noindent {\bf Word Representations.} Skip-Gram (SG) and
Continuous-Bag-of-Words (CBOW) \cite{mikolov2013efficient,mikolov2013distributed} are both local window context based word vectors. The former predicts the context words using the center word, while the latter predicts the center word using its
neighbors. \newcite{levy2014dependency} improve the word
embeddings by injecting syntactic information from the dependency parse trees. Fasttext \cite{bojanowski2017enriching} enriches word embeddings with subwords. Previous work also investigates how to incorporate external semantic knowledge into above word embeddings. \newcite{faruqui2015retrofitting} and \newcite{mrkvsic2016counter} propose to post-process word vectors using word synonymy or antonymy knowledge from semantic lexicons or task-specific ontology. \newcite{vashishth2019incorporating} leverage graph neural networks to incorporate syntactic and semantic knowledge into word embeddings. \newcite{alsuhaibani2018jointly} jointly train word vectors using a text corpora and a knowledge base, which contains special semantic relations.
All these methods inject semantic information from structured data, which is expensive to construct, while our methods can benefit from large-scale language models which is pre-trained on unlabelled corpora.

\noindent {\bf Distilling Knowledge From BERT.}
\newcite{hinton2015distilling} introduce knowledge distillation, which can transfer knowledge from a large (teacher) network to a small (student) network.
Following this idea, \newcite{tang2019distilling} propose to distill knowledge from the last layer of BERT into a single-layer BiLSTM network. \citet{sun2019patient} propose Patient-KD, which not only learns from the last layer of BERT, but also learns from multiple intermediate layers by two strategies: PKD-last and PKD-skip.
TinyBERT \cite{jiao2019tinybert} learns from the embedding layer, the hidden states and attention matrices of intermediate layers, and the logits output of the prediction layer of BERT.
\citet{Chen2020DistillingKL} improve sequence to sequence text generation models by leveraging BERT's bidirectional contextual knowledge. Different from all the previous work, in this paper, we present to distill semantic knowledge from a pre-trained  BERT into static word vectors (GloVe) using the weights of SAN and MLM.

\section{Background} \label{Background}
In this section, we briefly review GloVe and BERT, which our models are based on.
\subsection{GloVe} \label{background::glove}
Given a training corpus, GloVe first obtains the global word-word co-occurrence counts matrix $\mathbf{X}$, 
whose entries $\mathbf{X}_{ij}$ represents the total number of times word $w_j \in V$ occurring in the context of word $w_i \in V$, where $V$ is the word vocabulary of the training corpus. 
Formally, let $C(w_i)$ be the local window context of $w_i$. 
GloVe defines $\mathbf{X}_{ij}$ with regards to the position-based distance between $w_i$ and $w_j$ as:
\begin{equation}
    \mathbf{X}_{ij} = \sum_{w_j \in C(w_i)} \text{dis}(w_i, w_j) = \sum_{w_j \in C(w_i)} \dfrac{1}{|p_j-p_i|},
\label{eq:glovedistance}
\end{equation}
where $p_j$ and $p_i$ are positions in the context. Intuitively, words closer to $w_i$ receive larger weights. 

Denote the embedding of a target word $w_i$ and the context embedding of a context word $w_j$ as $\bm{e}_i$ and $\bm{e}'_j$, respectively. GloVe learns the word vectors by optimizing the following loss function:

\begin{equation}
J = \sum_{i=1}^{V} \sum_{j=1}^{V} \; f(\mathbf{X}_{ij}) \Big( \bm{e}_i^T \bm{e}'_j + b_i + b'_j - \log \mathbf{X}_{ij} \Big)^2,
\end{equation}
where $V$ is the size of vocabulary, $b_i$ and $b_j$ represent biases for $w_i$ and $w_j$, and $f(\cdot)$ is a weighting function. $f(\cdot)$ assigns lower weights to non-frequent co-occurrences:
\begin{equation}
 f(X_{ij}) = 
 \begin{cases} 
 (\mathbf{X}_{ij} / {x_{\text{max}}})^\alpha & \text{if } \mathbf{X}_{ij} < x_\text{max} \\ 
 1 &                                  \text{otherwise},
 \end{cases}
\end{equation}
where $x_{\text{max}}$ and $\alpha$ are hyper-parameters. 

After training with optimization methods, $\bm{e}_i + \bm{e}'_i$ is taken as the final word embeddings for $w_i$.

\subsection{BERT} \label{background::bert}
BERT is trained from large scale raw texts using masked language modeling task (MLM) with a deep bidirectional Transformer, which consists of multiple self-attention encoder (SAN) layers. Specifically, MLM masks a certain token as a special symbol $\langle \textsc{mask} \rangle$ (or a random token) randomly, and predicts the masked token using the contextualized output of the topmost layer.

Formally, given a token sequence $W = \{w_1,w_2,...,w_{n}\}$, a certain token $w_i$ ($i \in {[1...n]}$) is masked. The input layer $\mathbf{H}^0$  and each intermediate layer representation ${\mathbf{H}}^j$ are defined as:
\begin{equation}
  \begin{split}
    &\mathbf{H}^0 = [\bm{e}_1;...;\bm{e}_{i-1};\bm{e}_i;\bm{e}_{i+1};...;\bm{e}_n] + \mathbf{W}_p \\
    &{\mathbf{H}}^j = {\textbf{SAN}\_\textbf{Encoder}}({\mathbf{H}}^{j-1}) \ \ j \in [1...K],\\
  \end{split}
  \label{position encoding}
\end{equation}
where $\bm{e}_i$ is the embedding of $w_i$, $\mathbf{W}_p$ is the position embedding matrix, ${\textbf{SAN}\_\textbf{Encoder}}$ is SAN based encoder and $K$ denotes the number of SAN layers, respectively. More details about ${\textsc{San}\_\textsc{Encoder}}$ can be found in \cite{vaswani2017attention}.
For MLM, BERT predicts $w_i$ using
\begin{eqnarray} 
    &\text{logits}(\mathbf{h}^K_{i}) = \mathbf{W} \mathbf{h}^K_{i} \label{eq:background:logits}\\
    &\mathbf{p}[w_i] = \it \text{softmax}(\text{logits}(\mathbf{h}^K_{i}))
\label{eq:backgroud:softmax}
\end{eqnarray}
where $\mathbf{W}$ is a model parameter and $\mathbf{p}[w_i]$ denotes $P(w_i|w_1,...,w_{i-1}$, $\text{$\langle \textsc{mask} \rangle$}, w_{i+1},...,w_n)$.

Given a set of unlabelled text, $D=\{W^i\}|_{i=1}^N$, BERT is trained by maximizing the following objective function:
\begin{equation} 
    J=\sum_{i=1}^N \sum_{j=1}^{|W_i|} \mathbf{P}[w_i^j].
\label{eq:backgroud:glove:objective}
\end{equation}

\section{Semantic GloVe} 
\label{sec:method}
We replace the hard counts of co-occurrences into real values from BERT, according to self-attention scores and MLM probabilities, respectively.
\subsection{Semantic Co-occurrences from Multi-Head Self-Attention} 
\label{sec:san}
In this section, we leverage the multi-head self-attention weights of BERT to measure the semantic relationships of tokens instead of the heuristic position-based distance function of GloVe. 

Specifically, given a word sequence $W=\{w_1,...,w_K\}$, a window size $S$ and a pre-trained BERT model, 
we wish to define the word-to-word semantic distance (i.e., the $\emph{dis}$ function in Equation~\eqref{eq:glovedistance}) using the self-attention weights of BERT. 
Since BERT uses word pieces or byte-pair encodings (BPE; \cite{sennrich2016neural}) to segment words into BPE tokens, 
we firstly convert the original BPE-to-BPE attention weights to word-to-word attention weights.

Let the corresponding BPE token sequence is $T=\{t_1,...,t_L\}$, $N$ and $M$ be the number of layers and heads of the BERT model, and $\mathbf{AT}_{ij} \in \mathbb{R}^{L \times L}$ be the BPE token attention weights matrix of the $j$-th head in the $i$-th layer. 
We sum all heads and layers attention weights into one BPE-to-BPE attention weight matrix as follows:
\begin{equation}
    \mathbf{AT} = \sum_{i=1}^N \sum_{j=1}^M \mathbf{AT}_{ij}.
\end{equation}
Then, we generate the word-to-word attention weight matrix $\mathbf{AW} \in \mathbb{R}^{K \times K}$ by averaging the BPE-to-BPE attention weights. 
For $w_j$ within the local window context of word $w_i$,  $j \in [i - S, i + S] \cap j \neq i$, we denote attention weight from word $w_i$ to $w_j$ as $\mathbf{AW}_{ij}$ following:
\begin{equation}
    \mathbf{AW}_{ij} = \dfrac{1}{m \times n}\sum_{k={s_1}}^{s_m}\sum_{l={t_1}}^{t_n} \mathbf{AT}(k, l),
\end{equation}
where $\mathbf{AT}(k, l)$ is the attention weight from BPE token $t_k$ to $t_l$, $m$ and $n$ are the number of subwords of $w_i$ and $w_j$, respectively. 
To remove semantic irrelevant words, we sort $\mathbf{AW}_{ij}$ in descending order, 
and select the top-$S$ words as $w_i$'s context words $\mathcal{C}(w_i)$. 

Finally, the distance between the target word $w_i$ and the context word $w_j$ is calculated
by the following $\mathit{Division}$ distance function:
\begin{equation}
         \text{dis}(w_i, w_j) = \mathbf{AW}_{ij} /\mathbf{AW}_{i1}
         \label{Divide}.
\end{equation}
For the whole corpus, the global word-to-word co-occurrence count matrix is accumulated as:
\begin{equation}
    \mathbf{X}_{ij} = \sum_{w_j \in \mathcal{C}(w_i)} \text{dis}(w_i, w_j).
\end{equation}

We name the GloVe model training on this semantic word-to-word co-occurrence counts as $\text{SemGloVe}_{sd}$.

\subsection{Semantic Co-occurrences from Masked Language Model} 
\label{sec:mlm}
BERT pre-trained using deep bidirectional Transformers with the MLM task provides a dynamic way to define context for target word, 
which can avoid semantic irrelevant word pairs from a local window context. 
In this section, we introduce another method to distill semantic word-word co-occurrence counts by leveraging the MLM task of BERT. 

Formally, given a word sequence $W=\{w_1,...,w_K\}$, the corresponding BPE token sequence $T=\{t_1,...,t_L\}$, a window size $S$ 
and the topmost layer output $\textbf{h}_i^{K}$ of the ${\textbf{SAN}\_\textbf{Encoder}}$,
we define the context tokens $\mathcal{C}(w_i)$ using the output tokens of the MLM of BERT,
and calculate the word-to-word semantic distance (i.e., the \emph{dis} in Equation~\eqref{eq:glovedistance}) using the $\text{logits}$ in Equation~\eqref{eq:background:logits}.

\begin{figure*}[t]
    \centering
    \subfigure[Vector Dimension]{
    \begin{minipage}{5cm}
        \includegraphics[width=1.0\textwidth]{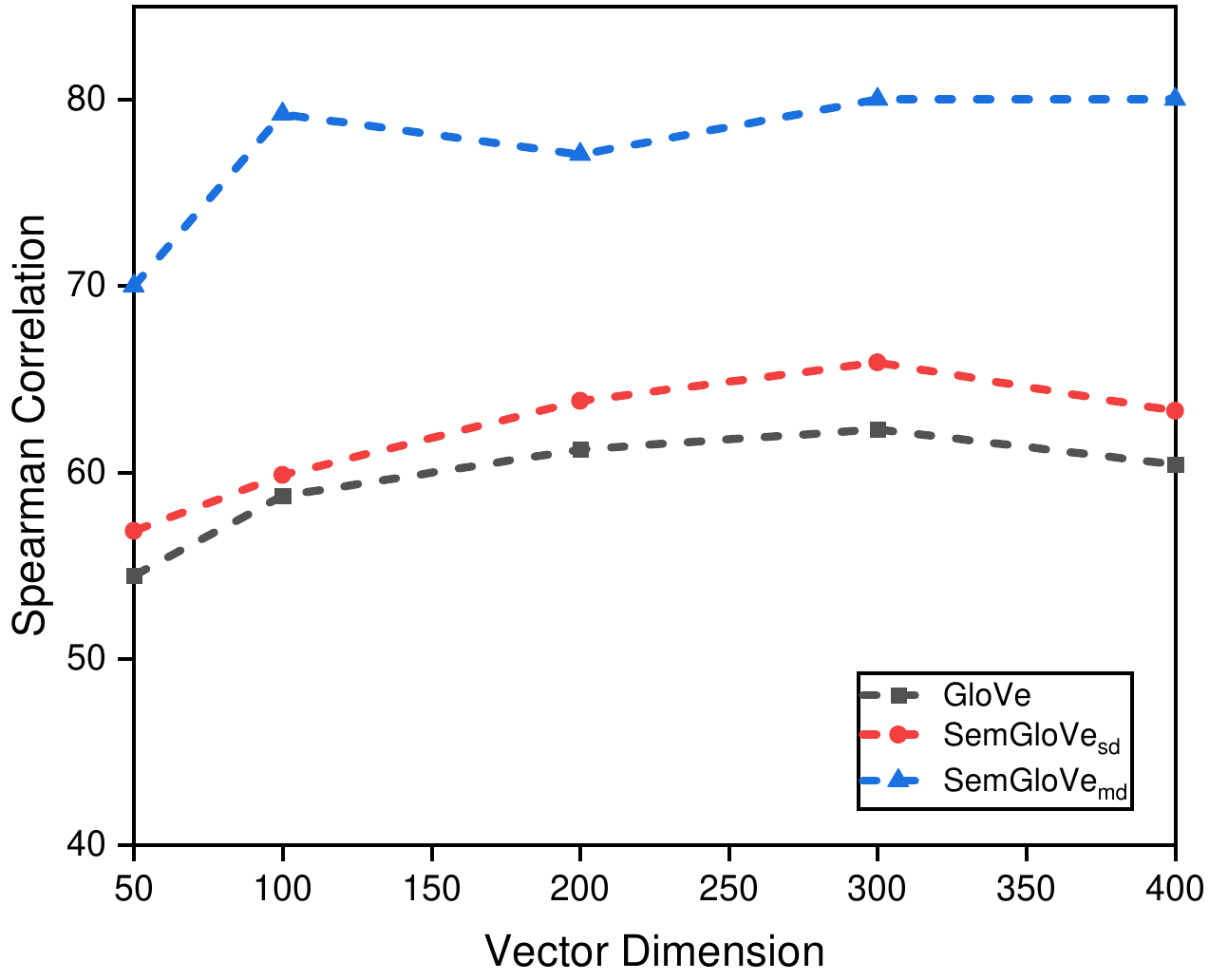}
		\label{fig:dev:dim}
    \end{minipage}
    }
    \subfigure[Corpus Size]{
    \begin{minipage}{5cm}
        \includegraphics[width=1.0\textwidth]{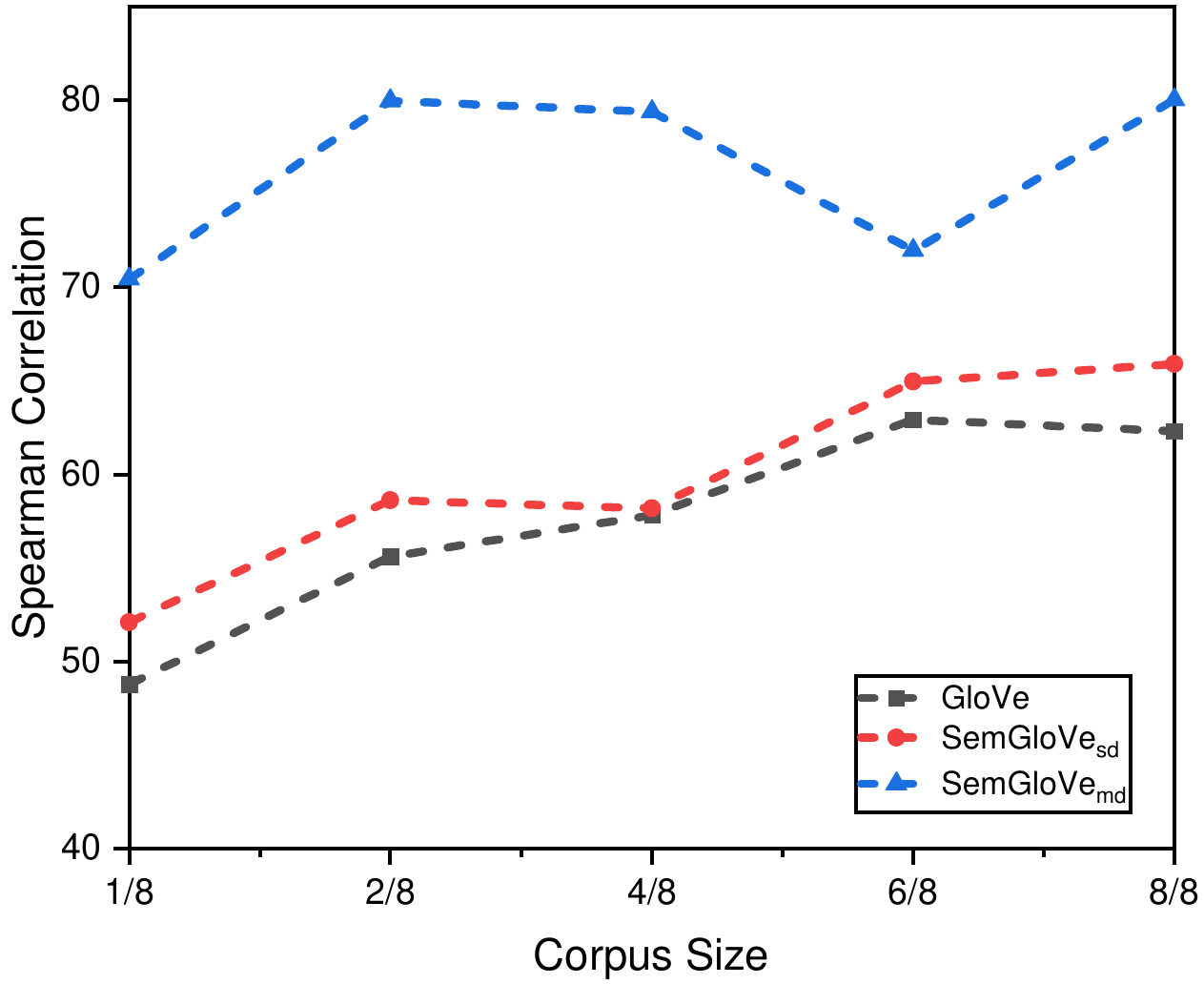}
        \label{fig:dev:corpussize}
    \end{minipage}
    }
    \subfigure[$X_\text{max}$ Value]{
    \begin{minipage}{5cm}
        \includegraphics[width=1.0\textwidth]{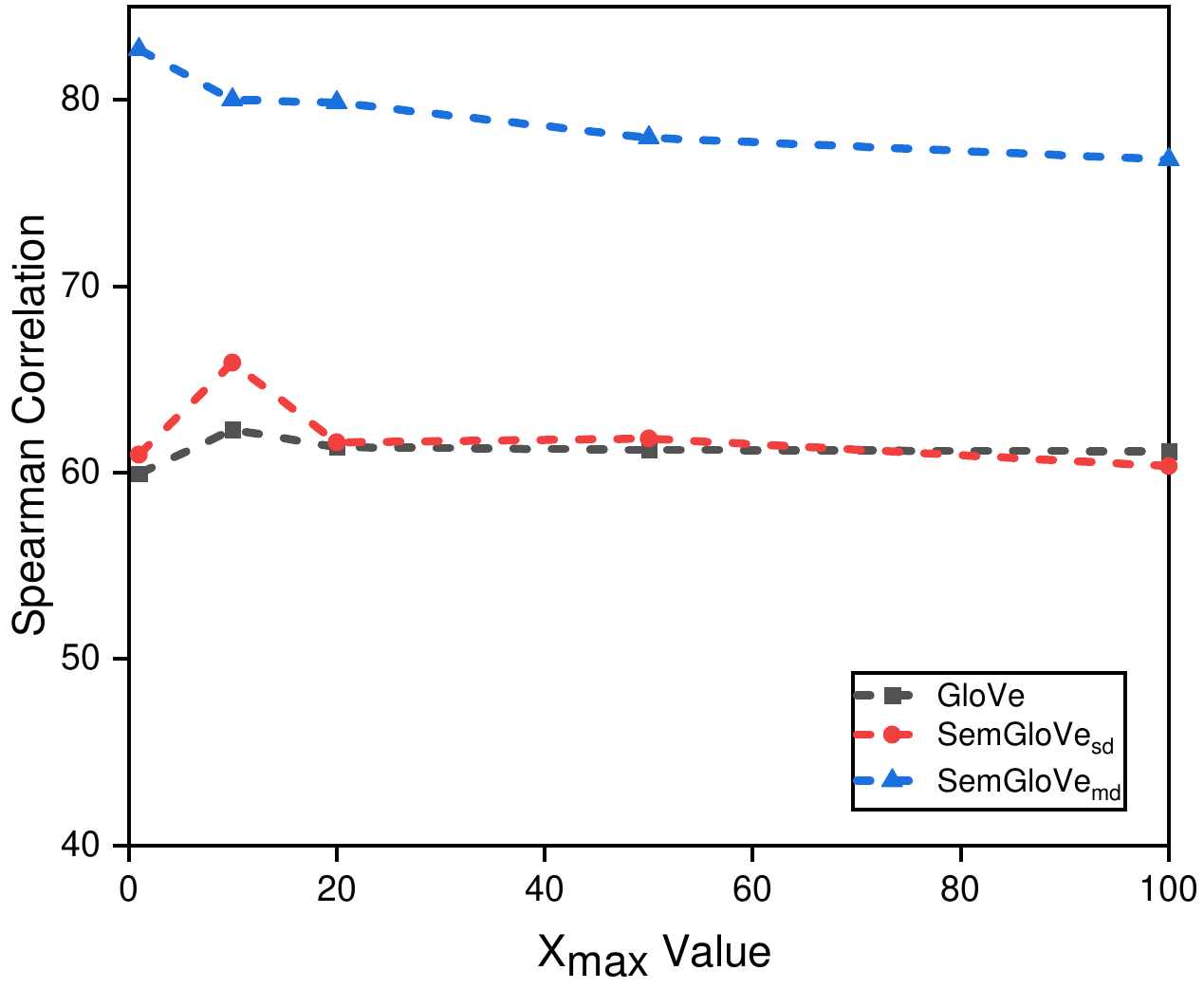}
        \label{fig:dev:xmax}
    \end{minipage}
    }
    \caption{Performance on WS353S dataset as function of vector dimension, corpus size and $x_\text{max}$ value.}
	\label{fig:dev:3figures}
\end{figure*}

Specifically,  
we first generate the BPE-to-BPE co-occurrence matrix $\mathbf{M} \in \mathbb{R}^{L \times L}$ to deal with the same BPE problem in Section~\ref{sec:san}.
To be more specific, we sort the output tokens of MLM in descending order with respect to $\text{logits}(\mathbf{h}^K_{i})$,
and then select the top $2S$ tokens to constitute $t_i$'s context tokens, denoted as $\mathcal{C}(t_i)=\{t_{i}^{(1)}, t_{i}^{(2)}, ..., t_{i}^{(2S)}\}$.
The corresponding logits are denoted as $\mathcal{G}(t_i)=\{g_{i}^{(1)}, g_{i}^{(2)}, ..., g_{i}^{(2S)}\}$.

Similar in Section \ref{sec:san}, the distance between the target token $t_i$ and the context token $t_{i}^{(j)}$ is calculated as:
\begin{equation}
\text{dis}(t_i, t_i^{(j)}) = g_{i}^{(j)} / g_{i}^{(1)}.
\end{equation}
Then, for the whole copus, the global BPE-BPE co-occurrence count from $t_i$ to $t_j$ is calculated as:
\begin{equation}
    \mathbf{M}_{ij} = \sum_{t_j \in C(t_i)} \text{dis}(t_i, t_j).
\end{equation}
Finally, we generate the global word-word co-occurrence matrix by averaging the BPE-to-BPE matrix as:
\begin{equation}
    \mathbf{X}_{ij} = \dfrac{1}{m \times n}\sum_{k={s_1}}^{s_m}\sum_{l={t_1}}^{t_n}\mathbf{M}_{kl},
\end{equation}
where $m$ and $n$ are the number of subwords of $w_i$ and $w_j$, respectively.

Ths GloVe model trained on this kind semantic co-occurrences is named as $\text{SemGloVe}_{md}$.

\section{Experiments}
We compare SemGloVe with GloVe on a range of intrinsic and extrinsic evaluation tasks, discussing the role the semantic co-occurrence plays in the training process.
\subsection{Settings}
We use the Wikipedia\footnote{https://dumps.wikimedia.org/enwiki/20180301} dump corpus as our training dataset, which is processed to only keep words appearing more than five times following \newcite{vashishth2019incorporating}. The dataset consists of 57 million sentences and 1.1 billion tokens.

For all our experiments, we set $x_{\text{max}}=10$, $\alpha=0.75$, the vectors dimension to 300, the window size $S$ to 5, and the number of iteration to 100. AdaGrad \cite{duchi2011adaptive} is used as the optimizer with initial learning rate $lr=0.05$. We dump the weights using the uncased BERT-large model with whole word masking, which has 24 layers and 12 attention heads\footnote{https://github.com/google-research/bert}.
\subsection{Baselines and Evaluation methods}
In addition to GloVe, we also compare SemGloVe with other state-of-the-art embedding methods, including Word2Vec \cite{mikolov2013distributed}, Deps \cite{levy2014dependency}, Fasttext \cite{bojanowski2017enriching}, SynGCN and SemGCN \cite{vashishth2019incorporating}, on several intrinsic and extrinsic semantic evaluation tasks. In addition to the GloVe baseline under the same settings, we also download GloVe$_\text{6B}$ from the official website as one baseline, which is trained on 6 billion tokens\footnote{https://nlp.stanford.edu/projects/glove/}.

We evaluate the intrinsic task on word similarity datasets, including WordSim-353 \cite{finkelstein2002placing},
and SimLex-999 \cite{kiela2015specializing}. Spearman correlation is taken as the main metric.

For extrinsic evaluation, we build a sentence-state LSTM (S-LSTM) \cite{zhang2018sentence} based sequence labeling model, which takes the concatenation of ELMo \cite{peters2018deep} and GloVe or SemGloVe as inputs. We compare this method with BERT$_\text{base}$ on a range of tasks, including Chunking, POS tagging and Named Entity Recognitoin (NER). The datasets used for evaluation are CoNLL-2000, the Wall Street Journal (WSJ) portion of the Penn Tree Bank (PTB) and CoNLL-2003, respectively.

\begin{table*}[t]
\centering
\small
\begin{tabular}{l|c|c|c|c|c}
\hline
\textbf{Models} & \textbf{WS353} & \textbf{WS353S} & \textbf{WS353R} & \textbf{SimLex999} & \textbf{Average}\\ \hline
Word2Vec           & 61.6 & 69.2 & 54.6 & 35.1 & 52.6 \\
Deps               & 60.6 & 65.7 & 36.2 & 39.6 & 50.5 \\
Fasttext           & 68.3 & 74.4 & 65.4 & 34.9 & 60.8 \\ 
SemGCN             & 60.9 & 65.9 & 60.3 & \textbf{48.8} & 59.0 \\
SynGCN             & 60.9 & 73.2 & 45.7 & 45.5 & 56.3 \\
\hline

GloVe              & 52.6 & 62.3 & 55.5 & 28.2 & 49.7 \\ 
GloVe$_\text{6B}$    & 56.5 & 64.6 & 50.0 & 31.3 & 50.6 \\
\hline

SemGloVe$_{sd}$      & 56.0 & 65.9 & 56.3 & 31.9 & 52.5 \\
SemGloVe$_{md}$      & \textbf{69.7} & \textbf{80.0} & \textbf{68.9} & 46.5 & \textbf{66.3} \\
\hline
\end{tabular}
\caption{Intrinsic Evaluation: Comparison on WordSim-353 and SimLex-999 datasets. The bold results are the best results.}
\label{tab:main:results}
\end{table*}

\subsection{Development Experiments}
Development experiments are conducted on the WS353S dataset to compare the performance of GloVe and SemGloVe with different hyper-parameters settings. To be specific, we investigate the influence of vector dimension, corpus size and $x_\text{max}$ value on the performance.

\paragraph{Effect of Vector Dimension.} We evaluate GloVe and SemGloVe with different vector dimensions, ranging from 50 to 400. For dimensions smaller than 300, the iteration number is set to 50, otherwise, it is set to 100. As shown in Figure \ref{fig:dev:3figures} (a), SemGloVe outperforms GloVe under different vector dimension settings.
When increasing the vector size from 50 to 300, both GloVe and SemGloVe can improve the performance on the WS353S dataset. However, when further increasing the vector size to 400, GloVe has a slight decrease in performance while SemGloVe$_{md}$ can keep the results stable. For a fair comparison, we choose 300 as the final vector size.

\paragraph{Effect of Corpus Size.} To investigate the influence of corpus size, we divide the whole training corpus into 8 parts. GloVe and SemGloVe are trained on 1/8, 2/8, 4/8, 6/8 and 8/8 corpus parts, respectively. According to the corpus size, $x_\text{max}$ is set to 1.25, 2.5, 5, 7.5 and 10, respectively. As shown in Figure \ref{fig:dev:3figures} (b), SemGloVe outperforms GloVe in all corpus size settings. As the corpus size increases, GloVe and SemGloVe$_{sd}$ obtain better performance. In the meantime, SemGloVe$_{md}$ can achieve reasonable results even with a small corpus size. We suppose that SemGloVe$_{md}$ can capture sufficient word-word co-occurrences counts under small corpus size, which we will analyze in Section \ref{analysis}.

\paragraph{Effect of $x_\text{max}$ Value.} To evaluate the effect of $x_\text{max}$, we train our models with different $x_\text{max}$ values. As shown in Figure \ref{fig:dev:3figures} (c), SemGloVe outperforms GloVe under different $x_\text{max}$ values. GloVe and SemGlove$_{sd}$ achieve the best results when $x_\text{max}$ is set to 10, while SemGloVe$_{md}$ maintains strong and stable results with different $x_\text{max}$ values. We suppose that, for window based contexts, a large $x_\text{max}$ value will lead to the neglect of useful word-word co-occurrences, while a small value will bring more noises. However, SemGloVe$_{md}$ can capture context from deep bidirectional Transformers, which leads to high quality co-occurrences. We set $x_\text{max}$ to 10 in the remaining experiments.
\begin{table}
\footnotesize
\begin{tabular}{l|c|c|c|c}
\hline
\textbf{Models} & \textbf{Chunking} & \textbf{POS} &  \textbf{NER} &
\textbf{Average} \\ \hline
BERT$_{base}$  &  96.60  & \textbf{97.77} & 92.40 & 95.59 \\
\hline
GloVe & 96.72 & 97.65 & 92.27 & 95.21 \\
\hline
SemGloVe & \textbf{96.75} & 97.70 & \textbf{92.60} & \textbf{95.68} \\
\hline
\end{tabular}
\caption{Extrinsic Evaluation: Comparison on Chunking, POS tagging and NER tasks. The bold results are the best results.} 
\label{tab:main:downstream}
\end{table}
\subsection{Final Results}
\paragraph{Intrinsic Evaluation Results.}
The final intrinsic evaluation results of SemGloVe and the baselines are listed in Table \ref{tab:main:results}. First, we find that SemGloVe$_{sd}$ and SemGloVe$_{md}$ outperform GloVe on the four word similarity evaluation datasets. Specifically, SemGloVe$_{sd}$ obtains 5.6\% absolute increase in performance on average, which demonstrates that self-attention weights of BERT are better to measure word similarities than the original position-based distance method.
In addition, compared with GloVe, SemGloVe$_{md}$ gives 33.4\% absolute increase in performance on average, which demonstrates that masked language model of BERT can model the context of words better than the predefined local window context, and produce more semantic relevant word pairs. Moreover, We also find that the averaged results of SemGloVe$_{md}$ outperform all the best methods in the literature. 

Since SemGloVe$_{md}$ performs better than SemGloVe$_{sd}$ on intrinsic tasks, we take them as the final SemGloVe.
\begin{table*}[t]
\small
\centering
\begin{tabular}{l|c|c|c|c|c}
\hline
\textbf{Models} & \textbf{WS353} & \textbf{WS353S} & \textbf{WS353R} & \textbf{SimLex999} & \textbf{Average}\\ \hline
GloVe              & 52.6 & 62.3 & 55.5 & 28.2 & 49.7 \\
\hline

SemGloVe$_{sr}$      & 54.0 & 62.8 & 54.9 & 29.3 & 50.3 \\
SemGloVe$_{sd}$      & 56.0 & 65.9 & 56.3 & 31.9 & 52.5 \\
\hline

SemGloVe$_{mr}$      & 67.8 & 79.2 & 66.7 & 45.1 & 64.7 \\
SemGloVe$_{md}$      & 69.7 & 80.0 & 68.9 & 46.5 & 66.3 \\

\hline
\end{tabular}
\caption{Comparisons of different distance function. The bold results are the best results.} 
\label{tab:main:distanc}
\end{table*}


\begin{table*}[t]
\centering
\small
\begin{tabular}{l|c|c|c|c|c}
\hline
\textbf{Models}& \textbf{WS353} & \textbf{WS353S} & \textbf{WS353R} & \textbf{SimLex999} & \textbf{Average} \\ \hline
GloVe(0.3B)                  & 52.6 & 62.3 & 55.5 & 28.2 & 49.7 \\ 
SemGloVe$_{sd}$(0.2B)   & 56.0 & 65.9 & 56.3 & 31.9 & 52.6 \\
SemGloVe$_{sd10}$(0.3B) & 55.3 & 63.2 & 56.4 & 28.6 & 50.9 \\
SemGloVe$_{sdR}$(0.2B)  & 52.7 & 60.5 & 54.6 & 27.7 & 48.9 \\
\hline
SemGloVe$_{md5}$      & 66.1 & 78.6 & 67.0 & 45.3 & 64.3 \\
SemGloVe$_{md}$      & 69.7 & 80.0 & 68.9 & 46.5 & 66.3 \\
\hline
\end{tabular}
\caption{Analysis of window size. The numbers in the parentheses are the total word pairs.} 
\label{tab:analysis:window}
\end{table*}

\paragraph{Extrinsic Evaluation Results.}
The final extrinsic evaluation results are shown in Table \ref{tab:main:downstream}.  We find that SemGloVe outperforms GloVe on three tasks, which demonstrates that SemGloVe contains more semantic information for downstream tasks. Specifically, SemGloVe gives 0.47\% absolute increase in performance on average, respectively.
Furthermore, compared with contextual word representations BERT$_\text{base}$, SemGloVe also shows competitive performance.

Note that the comparisons on extrinsic tasks are not to show the weakness of contextualized embeddings as input representations, but to show that SemGloVe (static word vectors) is still useful for downstream tasks. Another advantage of GloVe and SemGloVe, as compared with BERT, is that they are light weight and much faster for training and testing for downstream tasks.

\subsection{Analysis} \label{analysis}
We give analysis to the generated word-word co-occurrence counts of GloVe and SemGloVe, and try to answer the following research questions: (1) Can SAN weights of BERT be used to measure semantic similarity between word pairs, and lead to more accurate co-occurrence counts? (2) Can masked language model of BERT generate more semantic relevant word pairs? (3) Can the proposed \textit{Division} distance function be better than the original position-based distance function?

\noindent {\bf Co-occurrence Counts Analysis.} To answer above questions, we analyze the difference of co-occurrence counts between GloVe and SemGloVe 
by investigating some representative co-occurrence counts. As shown in Table \ref{tab:analysis:paircount}, word pairs of the first three rows, which are close in semantic, such as $<$paris, french$>$, can be found in all five methods. However, SemGlove$_{sd}$ can generate larger co-occurrence counts compared with GloVe, which proves that self-attention weights of BERT can assign larger counts to more semantic similar word pairs within the context. SemGloVe$_{md}$ can reach an average of 50 times counts of GloVe or SemGloVe$_{sd}$. We suppose that the words in a window context can be diverse and noisy, and thus distract the total co-occurrence counts. This helps to explain why SemGloVe$_{md}$ trained on small size corpus can still achieve strong performance. Furthermore, SemGloVe$_{md}$ can generate semantic relevant word pairs in the second three rows, which cannot be found in GloVe or SemGloVe$_{sd}$. This is because the two methods generate word pairs from a local window, while SemGloVe$_{md}$ can output most similar words over the whole vocabulary based on a deep bidirectional context.
\begin{figure*}[t] 
    \centering
    \subfigure[GloVe]{
        \begin{minipage}{5cm}
        \centering
        \includegraphics[width=1\textwidth]{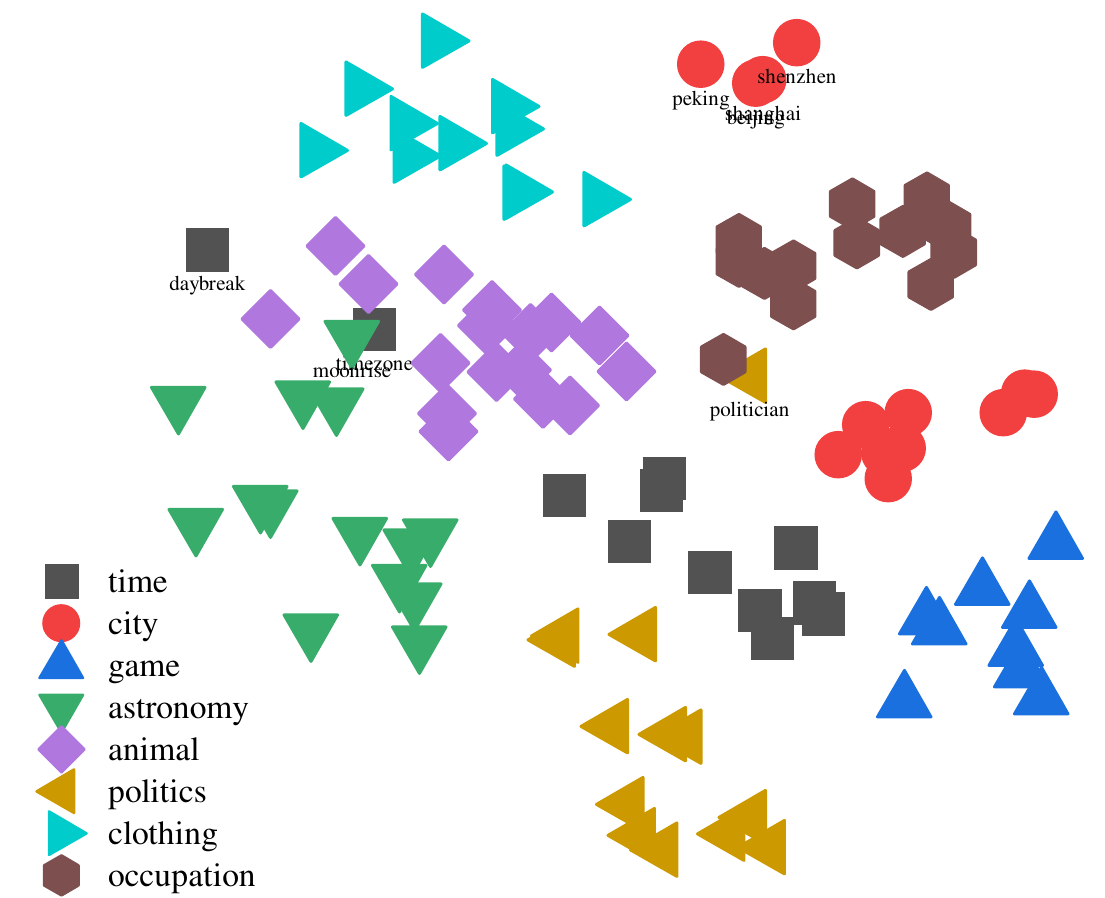}
        \end{minipage}
    }
    \subfigure[SemGloVe$_\text{sd}$]{
        \begin{minipage}{5cm}
        \centering
        \includegraphics[width=1\textwidth]{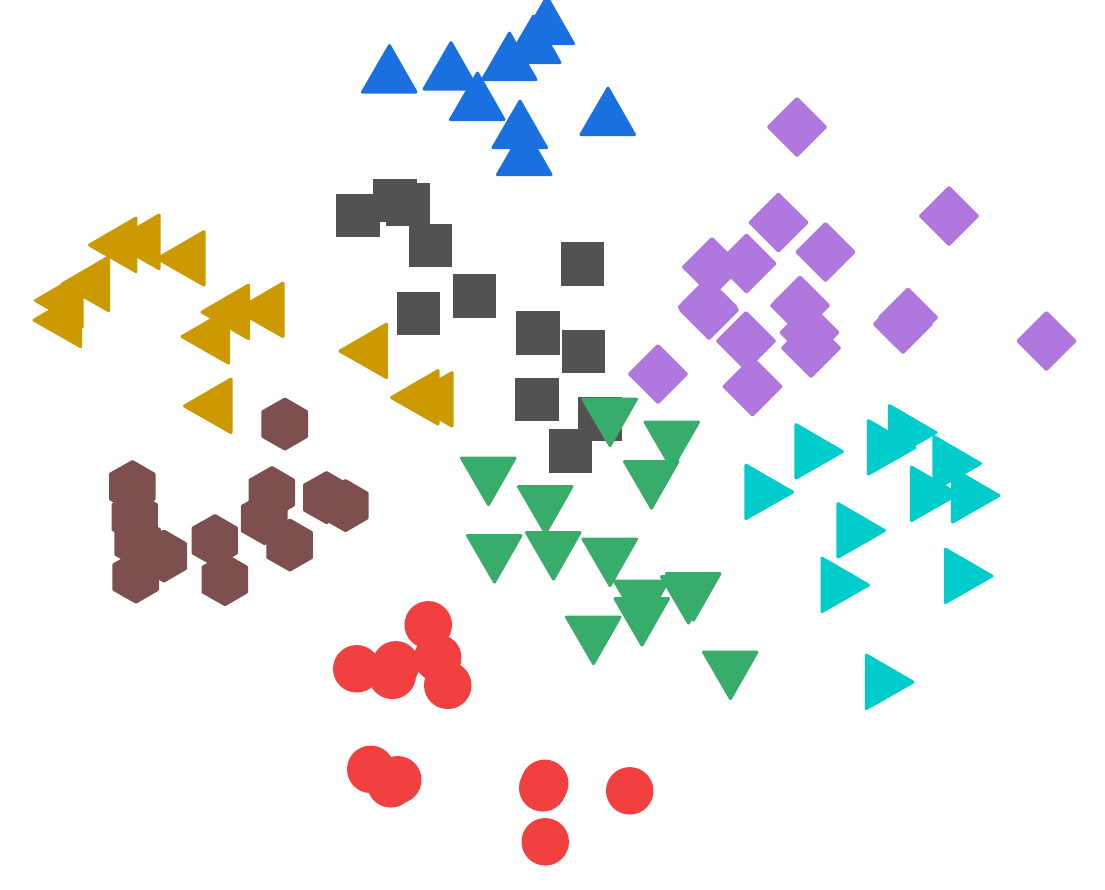}
        \end{minipage}
    }
    \subfigure[SemGloVe$_\text{md}$]{
        \begin{minipage}{5cm}
        \centering
        \includegraphics[width=1\textwidth]{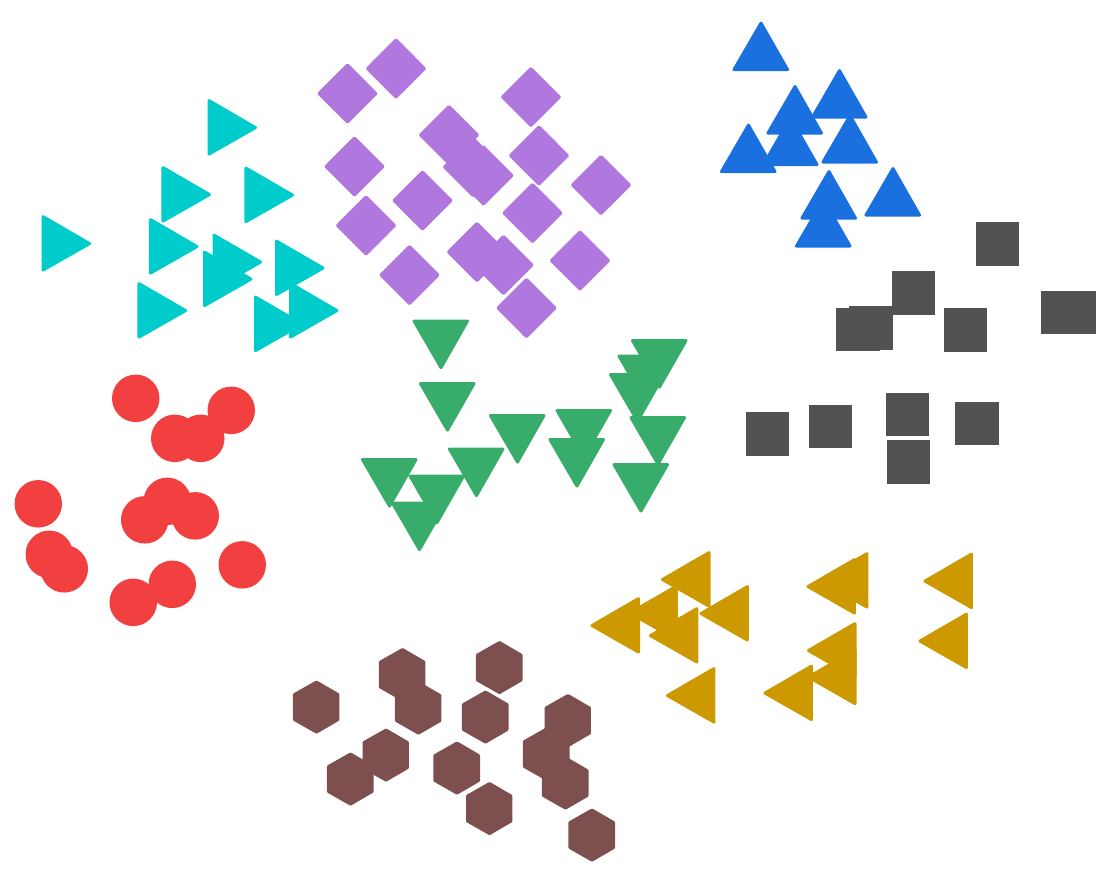}
        \end{minipage}
    }
	\caption{Visualization of three methods on randomly selected eight classes of words, including time, city, game, astronomy, animal, politics, clothing and creator. Each class contains ten words on average.}
	\label{figure:visualization}
\end{figure*}
\begin{table*}[t]
	\centering
	\small
	\begin{tabular}{l|c|c|c}
		\hline \textbf{Methods} &  GloVe & SemGloVe$_{sd}$ & SemGloVe$_{md}$ \\
		\hline
		$\langle$ \text{cat, cats} $\rangle$ & $0.33 \times 10^2$ & $0.86 \times 10^2$ & $3.48 \times 10^4$ \\
		$\langle$ \text{paris, french} $\rangle$  & $8.23 \times 10^2$ & $1.41 \times 10^3$ & $5.21 \times 10^4$ \\
		$\langle$ \text{man, woman} $\rangle$ & $1.48 \times 10^3$ & $2.57 \times 10^3$ & $1.25 \times 10^5$ \\ \hline 
		$\langle$ \text{himself, oneself} $\rangle$ & 0 & 0 & $4.42 \times 10^4$ \\
		$\langle$ \text{meanwhile, meantime} $\rangle$     & 0                  & 0                  & $2.72 \times 10^4$ \\
		$\langle$ \text{additionally, moreover} $\rangle$  & 0                  & 0                  & $2.90 \times 10^4$ \\ 
		\hline 
		$\langle$ \text{biophysical, biochemical} $\rangle$ & 12.53               & 12.93               & $1.16 \times 10^2$ \\
		$\langle$ \text{egyptologist, archaeologist} $\rangle$ & 15.00               & 17.30               & $0.39 \times 10^2$ \\
		$\langle$ \text{egyptologist, egyptian} $\rangle$ & 11.32               & 8.89               & $8.58 \times 10^3$ \\
		\hline
	\end{tabular}
	\caption{Some specific word-word co-occurrence counts of GloVe and SemGloVe.}
	\label{tab:analysis:paircount}    
\end{table*}

SemGloVe$_{md}$ can also generate rich semantic word embeddings because of averaging counts of the subword pairs. We give some examples in the third three rows of Table \ref{tab:analysis:paircount}. For instance, word ``egyptologist" consists of two subwords ``egypt" and ``\#\#ologist", which lead to rich semantic word pairs, $<$egyptologist, egyptian$>$ and $<$egyptologist, archaeologist$>$, respectively.


\noindent \textbf{Distance Function.}
Different from the position-based distance function of GloVe, our \textit{Division} distance function is based on either the weights of SAN or the output logits of MLM. To investigate whether our distance function captures more semantic knowledge, we first sort the context words according to their SAN weights or MLM logits in descending order, and then directly replace the \textit{Division} function of SemGloVe with position-based distance function, and name the two corresponding method as SemGloVe$_{sr}$ and SemGloVe$_{mr}$. The comparisons are listed in Table \ref{tab:main:distanc}. We can observe that the \textit{Division} distance function performs better than the original distance function of GloVe in mapping BERT's weights into co-occurrence counts, which can be because that the $\textit{Division}$ function leverage information of weights better than the original function.

\noindent \textbf{Visualization.}
We use t-SNE \cite{maaten2008visualizing} to visualize GloVe, SemGloVe$_{sd}$ and SemGloVe$_{md}$ embeddings. As shown in Figure \ref{figure:visualization}, GloVe has several outliers for classes of city, time, astronomy and politics. One reason is that for outlier words, such as ``daybreak", there are few usefull related word pairs in the co-occurrence matrix of GloVe. SemGloVe$_{sd}$ performs better than GloVe, and has fewer outliter words. However, some classes are not seperated clearly. In contrast, SemGloVe$_{md}$ gives the best visualization result among all models. Words from the same class are clustered together, and different classes have clear boundaries. We attribute this to the rich semantic word pairs of SemGloVe$_{md}$. For the same word ``daybreak", the co-occurrence matrix of SemGloVe$_{md}$ instead contains semantic relevant word pairs, such as $<$daybreak, today, $1.54 \times 10^3$ $>$, $<$daybreak, sunday, $3.20 \times 10^3$$>$.

\noindent \textbf{Window Size Analysis.} Although SemGloVe with SAN and GloVe have the same window size, the context words in GloVe are twice as many as those in SemGloVe. This is because the context of GloVe is symmetric. In SemGloVe with SAN, in fact, we sort the scores of all the 10 words in descending order, and then select the top 5 words as context words. To analyze the effect of window size, we take ablation studies on the word similarity dataset as shown in Table \ref{tab:analysis:window}. ``SemGloVe$_{sd10}$" is a setting, where we select the top 10 words as context words, while for ``SemGloVe$_{sdR}$", we replace the co-occurrence counts of ``SemGloVe$_{sd}$" with the corresponding counts of GloVe. We can find that the performance of ``SemGloVe$_{sd10}$" is slightly worse than ``SemGloVe$_{sd}$", but still better than GloVe. However,  ``SemGloVe$_{sdR}$" reaches almost the same results as GloVe. We conclude that reducing the window size in SemGloVe$_{sd}$ can reserve valuable word pairs.

We also reduce the window size from 10 to 5 in SemGloVe with MLM, and name this setting as SemGloVe$_{md5}$. As shown in Table \ref{tab:analysis:window}, we can find that reducing the window size in MLM model will decrease the performance.

\section{Conclusion}
We propose SemGloVe, which replaces the hard counts of GloVe by distilling semantic co-occurrences from BERT. Compared with GloVe, SemGloVe can extract word pairs without local window constraints, and can count co-occurrences by directly considering the semantic distance between word pairs. Intrinsic and extrinsic experiments show that SemGloVe outperforms GloVe.

\bibliography{acl2020}
\bibliographystyle{acl_natbib.bst}
\end{document}